\documentclass[journal, 10pt, letter]{IEEEtran}
\usepackage{amsmath,amssymb,amsfonts,bbm,bm}
\usepackage{color}
\usepackage{multirow}
\usepackage{graphicx}
\usepackage[linesnumbered,ruled]{algorithm2e}
\usepackage{booktabs}
\usepackage{mwe}
\usepackage{cite}
\usepackage{graphicx}
\usepackage{textcomp}
\usepackage[normalem]{ulem}
\usepackage{xcolor}
\usepackage{tabulary}
\usepackage{array} 
\usepackage{hyperref}
\usepackage{algpseudocode}
\newcommand{\thetab}{\ensuremath{\bm{\theta}}}
\newcommand{\phis}{\ensuremath{\bm{\phi}}}
\newcommand{\BfPara}[1]{{\noindent\bf#1.}\xspace}
\begin{document}

\title{Slimmable Quantum Federated Learning}
\author{
$^\dag${Won Joon Yun},
$^\dag${Jae Pyoung Kim},
$^\ddag${Soyi Jung},
$^\circ${Jihong Park},
$^\S${Mehdi Bennis}, and
$^\dag${Joongheon Kim}
\\
\IEEEauthorblockA{$^\dag$School of Electrical Engineering, Korea University, Seoul, Republic of Korea}
\IEEEauthorblockA{$^\S$Centre for Wireless Communications, University of Oulu, Oulu, Finland}
\IEEEauthorblockA{$^\ddag$School of Software, Hallym University, Chuncheon, Republic of Korea}
\IEEEauthorblockA{$^\circ$School of Information Technology, Deakin University, Geelong, Victoria, Australia}
\thanks{*Corresponding author: S. Jung, J. Park and J. Kim are the corresponding author of this paper.}
\thanks{E-mail: \texttt{sjung@hallym.ac.kr}, \texttt{{jihong.park@deakin.edu.au}}, \texttt{{joongheon@korea.ac.kr}}}
\thanks{This research was supported by NRF-Korea (2021R1A4A1030775, 2022R1A2C2004869, 2019M3E4A1080391) and also by MSIT (Ministry of Science and ICT), Korea, under the ITRC (Information Technology Research Center) support program (IITP-2022-2017-0-01637) by IITP (Institute for Information \& Communications Technology Planning \& Evaluation)}
}

\maketitle

\begin{abstract}
Quantum federated learning (QFL) has recently received increasing attention, where quantum neural networks (QNNs) are integrated into federated learning (FL). In contrast to the existing static QFL methods, we propose \emph{slimmable QFL (SlimQFL)} in this article, which is a dynamic QFL framework that can cope with time-varying communication channels and computing energy limitations. This is made viable by leveraging the unique nature of a QNN where its angle parameters and pole parameters can be separately trained and dynamically exploited. Simulation results corroborate that SlimQFL achieves higher classification accuracy than Vanilla QFL, particularly under poor channel conditions on average.
\end{abstract} 

\section{Introduction}\label{sec:1}
Recent advances in noisy intermediate-scale quantum (NISQ) computing processors \cite{arute2019quantum} and machine learning (ML) algorithms \cite{burkart2021survey} appear to be a prelude to the era of quantum ML (QML) \cite{schuld2022quantum}. Just like the neural network (NN) of classical ML, QML is implemented by a quantum neural network (QNN) in which a parameterized quantum circuit (PQC) adjusts the input quantum qubit states, and the expected measurement determines the output for a given basis \cite{bharti2022noisy}. With fewer parameters, QML has achieved the level of performance comparable to classical ML in classification \cite{schuld2021supervised,havlivcek2019supervised}, data generation \cite{zoufal2019quantum}, and reinforcement learning tasks \cite{jerbi2021variational}. Meanwhile, by combining federated learning (FL) \cite{chen2021federated,huang2022quantum} with QML, quantum FL (QFL) has shown its potential in utilizing distributed data and quantum computing resources \cite{chehimi2022quantum}. 

While interesting, the current QFL, hereafter referred to as Vanilla QFL, is nothing more than iteratively averaging the PQC parameters. Therefore, it is difficult to cope with environmental dynamics such as time-varying communication channel conditions and energy limitations~\cite{matsubara2022bottlefit}. To overcome this limitation, we propose a novel dynamic QFL framework, coined \emph{slimmable QFL (SlimQFL)}, inspired by the resemblance between the slimmable NN (SNN) architecture \cite{infocom2022baek} and a dyadic nature of the QNN architecture as elaborated next.

\begin{figure}[t!]\centering
\includegraphics[width=1.\linewidth]{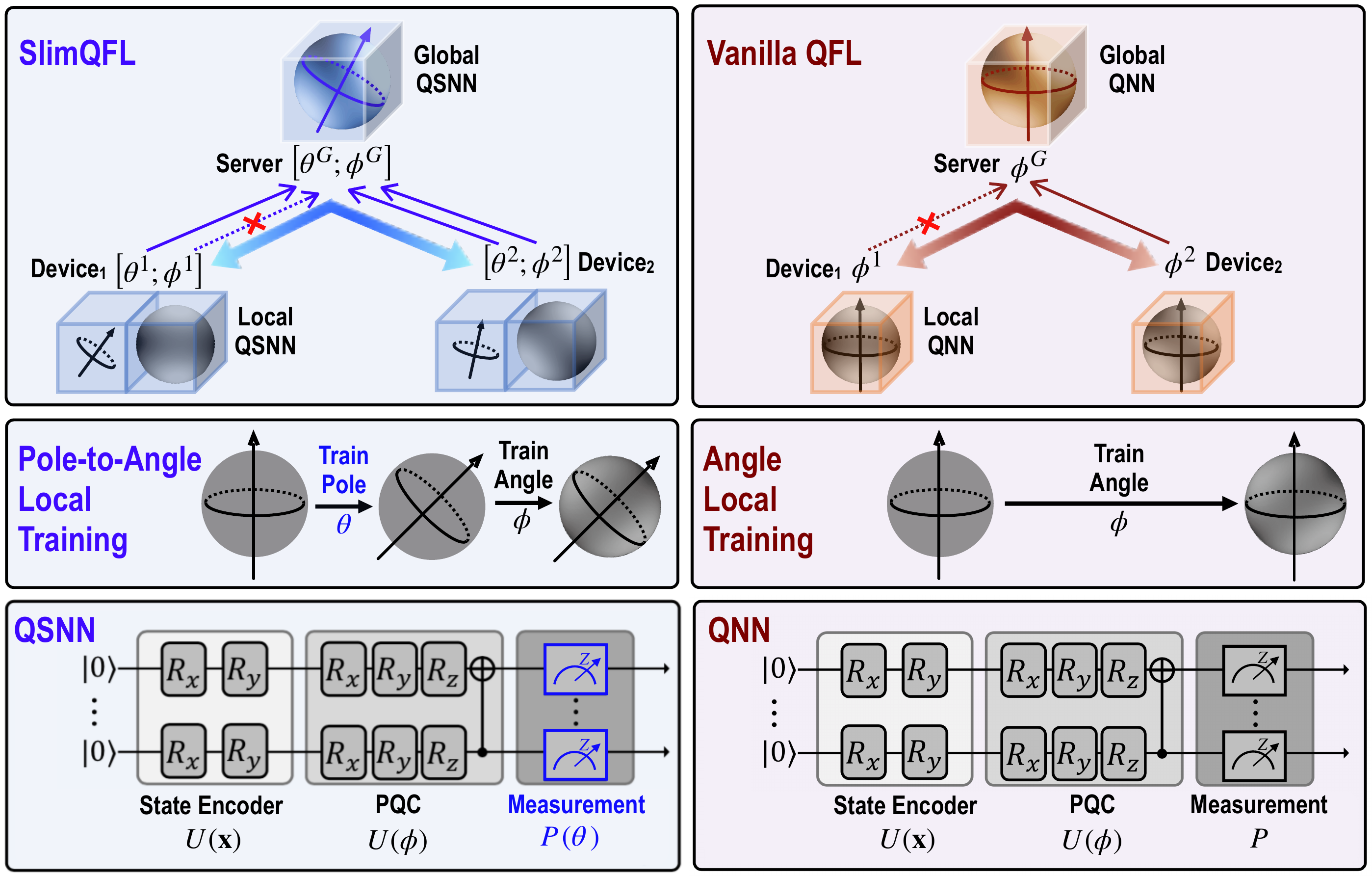}\\
\caption{A schematic illustration of SlimQFL (left) wherein the pole and angle parameters of local QSNN models can be separately trained and dynamically communicated, in contrast to Vanilla QFL (right) based on QNNs wherein only angle parameters are trainable.}
\label{fig:abstract}
\end{figure}

The SNN of classical ML is a dynamic architecture in the sense that not only can the entire parameters be trained, but also a fraction of the parameters can be separately trained and exploited. Slimmable FL (SlimFL) utilizes such SNNs as the local models of devices, thereby responding to the time-varying energy and communication channel conditions \cite{slimfl2022yun}. Similarly, a QNN can be viewed as two sets of separately tunable parameters: angle parameters of the PQC and pole parameters of the measurement basis. While existing QNN architectures and training algorithms focus only on the angle parameters \cite{chen2021federated}, we propose a quantum SNN (QSNN) wherein both angle and pole parameters can be separately trainable for SlimQFL. 

Consequently, during the local training of SlimQFL, each device first trains the pole parameters of its local QSNN, followed by the angle parameters, henceforth referred to as pole training and angle training, respectively. Then, depending on the channel condition during a communication round, each device transmits either the both angle and pole parameters or only the pole parameters ($7$x smaller in our experiments). A server produces the averaged local QSNNs as a global QSNN that is downloaded by each device. Numerical experiments corroborate that SlimQFL achieves {11.7\%} higher accuracy than Vanilla QFL in MNIST classification, thanks to its successful reception of at least pole parameters even under poor channel conditions. We additionally validate that pole training without angle training can indeed improve accuracy, demonstrating the potential of QSNN as a dynamic QNN architecture.

\section{Preliminaries: From Quantum Machine Learning to Quantum Federated Learning}
\BfPara{Basic Quantum Gates} A qubit is a quantum computing unit where the quantum state is represented with two basis $|0\rangle ,|1\rangle$ in Bloch sphere~\cite{bouwmeester2000physics}. The quantum state is written as $|\psi\rangle= \alpha|0\rangle + \beta|1\rangle$, where $\alpha^2+\beta^2=1$. Suppose there is a single qubit system, a classical data $\delta$ is encoded to quantum state with the rotation gates $R_x(\delta)$, $R_y(\delta)$, and $R_z(\delta)$, where $R_x(\delta)$, $R_y(\delta)$, and $R_z(\delta)$ represent the rotation of $\delta$ over $x$-, $y$-, and $z$-axes in Bloch sphere, respectively. In a multiple qubit system, qubits can be entangled with \textit{controlled-NOT (CNOT)} gates. These basic quantum gates configure the QNNs.

\BfPara{QNN based QML} 
As shown in Fig.~\ref{fig:abstract}(right), the structure of a QNN is tripartite: the state encoder, PQC, and the measurement layer~\cite{killoran2019continuous}. 
In the forward propagation, classical input data $\mathbf{x}$ needs to be first encoded with the state encoder via basic rotation gates, which is a unitary operation and denoted as $U(\mathbf{x})$. 
Then, the encoded quantum state is processed through the PQC $U(\phis)$, a multi-layered set of CNOT gates and rotation gates associated with trainable parameters $\phis$. 
The output of the PQC is the entangled quantum state that can be measured after applying a projection matrix $P$ onto the reference $z$-axis. 
The measured output $\langle O \rangle_{\mathbf{x},\phis} \in [-1,1]^{|y|}$ is called an \textit{observable}, where $|y|$ denotes the output dimension. 
Given the observable and the ground-truth of input, the loss $\mathcal{L}(\phis)$ is calculated. 
Subsequently, the QNN is trained accordingly using the stochastic gradient descent algorithm:
\begin{align}
    \tilde \phis \leftarrow \phis - \eta \nabla_{\phis}\mathcal{L}(\phis) \label{Eq:local}
\end{align}
where $\eta$ is the learning rate, and the gradient $\nabla_{\phis}\mathcal{L}(\phis)$ is calculated using the parameter shift rule~\cite{mitarai18}. 

\BfPara{Vanilla QFL}
\texttt{FedAvg} is the standard algorithm in FL with classical NNs \cite{mcmahan2017communication}. In each communication round, the operations of \texttt{FedAvg} can be summarized by: (i) each device's local training of an NN; (ii) the parameter server's construction of a global NN by averaging the local NNs; and (iii) each device's replacement of its local NN with the global NN. In Vanilla QFL, (i) is implemented using \eqref{Eq:local}, and the global QNN $\tilde \phis^G$ in (ii) is given by the averaged PQC angle parameter, i.e., 
\begin{align}
    \tilde \phis^G \leftarrow \frac{1}{\sum^N_{n=1}c_n} \sum^N_{n=1} c_n \cdot \tilde \phis^n,
\end{align}
where $\tilde \phis^n$ is the $n$-th device's local NN, and $c_n \in \{0,1\}$ is an indicator function returning $1$ if the $n$-th device contributes to the global model aggregation. This principle is applied for a privacy-preserving application \cite{li2021quantum}, binary classification~\cite{chen2021federated,huang2022quantum}, and image classification tasks \cite{chen2021federated}. 

\section{Slimmable Quantum Federated Learning}
\begin{algorithm}
\caption{Pole-to-Angle Local-QSNN Train}\label{alg:a2p}
\scriptsize
\textbf{Notation.} $\mathcal{D}$: local trainset, $\mathbf{x}_i$: data of $i$-th batch, $y_i$: label of $i$-th batch, $\eta_l$: learning rate in $l$-th iterations.\;  
\textbf{Initialization.} local-QNN parameters, $\phis, \thetab$\;
 \For{ $l = \{1,2,\dots, L\}$}
 {
     \For{ $(\mathbf{x}_i, y_i) \in \mathcal{D}$}
             { 
                $\hat{y}_i \leftarrow \textsf{QSNN}(\mathbf{x}_i;\phis,\thetab)$ // $\hat{y}_i$: logits\;
                Calculate loss, $\mathcal{L}( \phis,\thetab, (\mathbf{x}_i, y_i))$\;
                Update pole, $\thetab \leftarrow \thetab-\eta_l\nabla_{\thetab}\mathcal{L}( \phis,\thetab, (\mathbf{x}_i, y_i))$\;
             }
 }
 $\tilde{\thetab} \leftarrow \thetab$ \;
 \For{ $l = \{1,2,\dots, L\}$}
 {
     \For{ $(\mathbf{x}_i, y_i) \in \mathcal{D}$}
             { 
                $\hat{y}_i \leftarrow \textsf{QSNN}(\mathbf{x}_i;\phis,\thetab)$\;
                Calculate loss, $\mathcal{L}( \phis,\thetab, (\mathbf{x}_i, y_i))$\;
                Update angle,$\phis \leftarrow \phis-\eta_l\nabla_{\phis}\mathcal{L}( \phis,\tilde \thetab, (\mathbf{x}_i, y_i))$\;
             }
 }
 $\tilde{\phis} \leftarrow \phis$\;
\end{algorithm}

Departing from Vanilla QFL based on QNNs, we aim to make QFL that can cope with dynamic environments such as time-varying communication channels and energy limitations. To this end, we propose a novel QSNN architecture, and develop SlimQFL involving local QSNN training and global QSNN aggregation operations, as elaborated next.

\BfPara{QSNN Architecture}
In a QNN, its feature map is trained by adjusting the PQC angle parameters \cite{havlivcek2019supervised}, and the corresponding qubit states can be represented on the Bloch sphere, as illustrated in Fig.~\ref{fig:abstract}(left). Meanwhile, the QNN output is given by the measured qubit states projected onto a hyperplane for a given basis pole of the Bloch sphere. While the standard QML focuses only on tuning the PQC angle parameters, (Schuld et al., 2022) show that it is possible to adjust the hyperplane by changing the pole of the measurement~\cite{schuld2022quantum}. Inspired by this, we propose a \emph{QSNN} where not only can the PQC angle parameters $\phis$ be trained but also the measurement pole parameters $\thetab$ can be trained. Therefore, the projection matrix $P_{\thetab}$ of QSNN is trainable in comparison to the fixed projection matrix $P$ in QNN.

\BfPara{Pole-to-Angle Local Training}
Like Vanilla QFL associated with local QNNs, each device under SlimQFL stores a local QSNN that is trained using its own local dataset. The new key element is that SlimQFL trains the angle and pole parameters of QSNN separately in a sequential way. Since the number of pole parameters is commonly smaller than that of angle parameters, we first train the pole parameters $\thetab$, then train the angle parameters $\phis$. After $L$ local iterations, the local QSNN $
[\tilde \thetab^n ; \tilde \phis^n]$ of the $n$-th device is updated as:
\begin{align}
\begin{bmatrix}
\tilde \thetab^n \\ \tilde \phis^n
\end{bmatrix}
\leftarrow 
\begin{bmatrix}
\thetab^n \\  \phis^n
\end{bmatrix}
- \eta_t
\begin{bmatrix}
\sum^L_{l=1} \nabla_{\thetab^n_l}\mathcal{L} (\phis^n,\thetab^n_l) \\
 \sum^L_{l=1} \nabla_{\phis^n_l}\mathcal{L}(\phis^n_l,\tilde \thetab^n)
\end{bmatrix},
\end{align}
 where $\eta_t$ denotes the learning rate at time $t$. 
 
\begin{algorithm}[t]
\caption{SlimQFL}\label{alg:sqfl}
\scriptsize
\textbf{Notation.} $\tilde{\thetab}^n,\tilde{\phis}^n$: $n$-th device's pole/angle parameters, $\tilde{\thetab}^G,\tilde{\phis}^G$: pole/angle parameters of server-side QSNN\;
\textbf{Initialization.} $\forall c^n_{\thetab},c^n_{\phis} \leftarrow 0$\;
 \For{ $n = \{1, \dots, N\}$}
 {
 Sample $\chi^n \sim \exp(1) $\;
     \If{$R^n \geq u^{\text{whole}}_{\text{th}}$}
             { 
             Transmit pole/angle parameters, $( \tilde{\thetab}^n,\tilde{\phis}^n)$\;
             $c^n_{\textit{\thetab}}, c^n_{\textit{\phis}}\leftarrow 1$ \;
             }
     \ElseIf{$R^n \geq u^{\text{pole}}_{\text{th}}$}
     {
             Transmit angle parameters, $\tilde{\thetab}^n$\; $c^n_{\thetab} \leftarrow  1$\;
     }
     
 } 
 \If{ $\sum^N_{n=1} c^n_{\thetab} \neq 0$ \text{and} $\sum^N_{n=1} c^n_{\phis} \neq 0$}
 {
 Update with~\eqref{eq:slimqfl}\;
 }
\end{algorithm}

\BfPara{Dynamic Global Model Aggregation} 
By leveraging the QSNN architecture, at each communication round, the $n$-th device uploads either: (i) only the pole parameters $\tilde \thetab^n$ or (ii) both the pole and angle parameters $[\tilde \thetab^n; \tilde \phis^n]$, depending on its communication channel condition, energy availability, and/or other time-varying environmental factors. The parameter server aggregates the uploaded parameters accordingly and constructs a global QSNN $[\tilde \thetab^G;\tilde \phis^G]$:
\begin{align} \label{eq:slimqfl}
\begin{bmatrix}
\tilde \thetab^G
 \\ \tilde \phis^G
\end{bmatrix}
\leftarrow 
\begin{bmatrix}
 \frac{1}{\sum^N_{n=1}c^n_{\thetab}} {\sum^N_{n=1}c^n_{\thetab}\tilde\thetab^n}\\
 \frac{1}{\sum^N_{n=1}c^n_{\phis}} {\sum^N_{n=1}c^n_{\phis}\tilde\phis^n}
\end{bmatrix},
\end{align}
where the indicator functions $c^n_{\thetab}$ and $c^n_{\phis}$ count pole and angle uploading events respectively. Finally, each device downloads the global QSNN $[\tilde \thetab^G;\tilde \phis^G]$, and iterates the operations mentioned above until convergence, as summarized in Algorithm~\ref{alg:sqfl}.

\section{Numerical Experiments}
\subsection{Simulation Settings} 
\begin{figure*}[t!]
   \noindent\begin{minipage}{0.33333\textwidth}
     \centering
\includegraphics[width=1\linewidth]{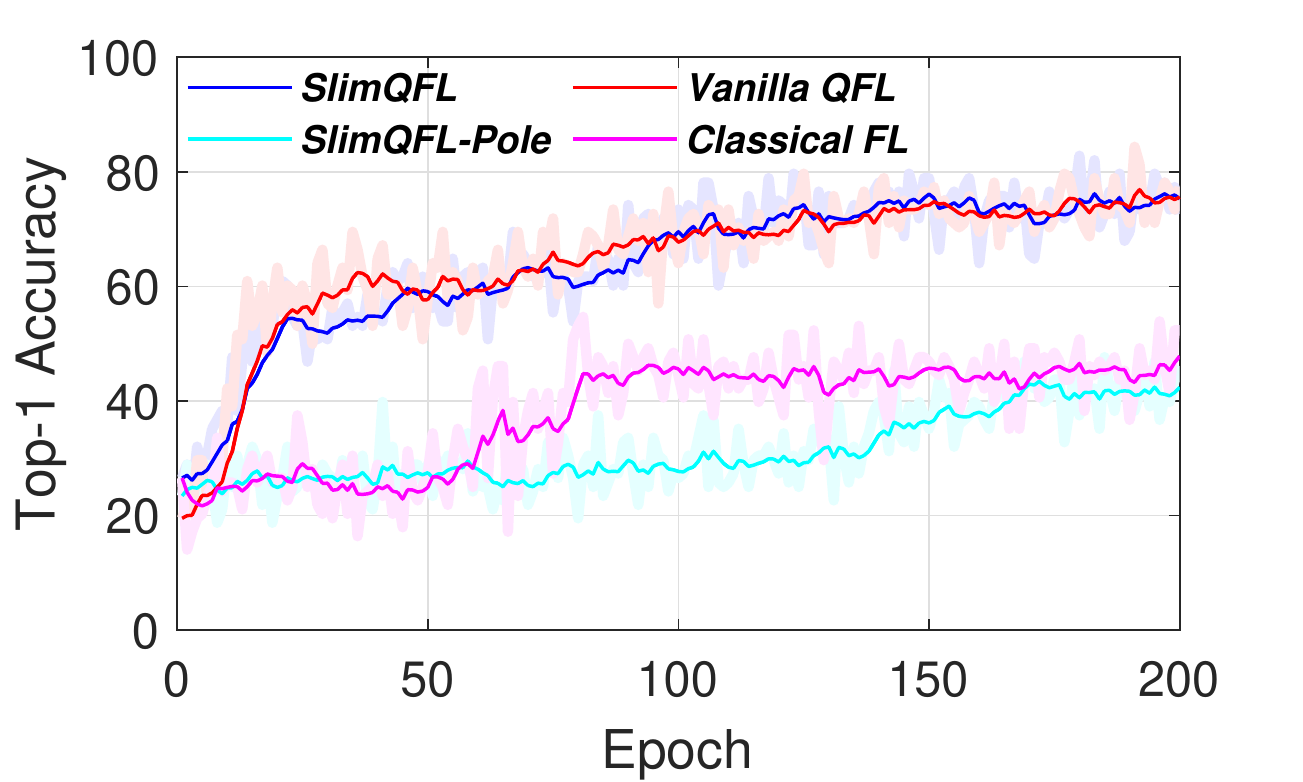}\\
\caption{ Good channel ($\sigma^2=-40$dB).}
    \label{fig:result-good}
    \end{minipage}\hfill
       \noindent\begin{minipage}{0.33333\textwidth}
     \centering
\includegraphics[width=1\linewidth]{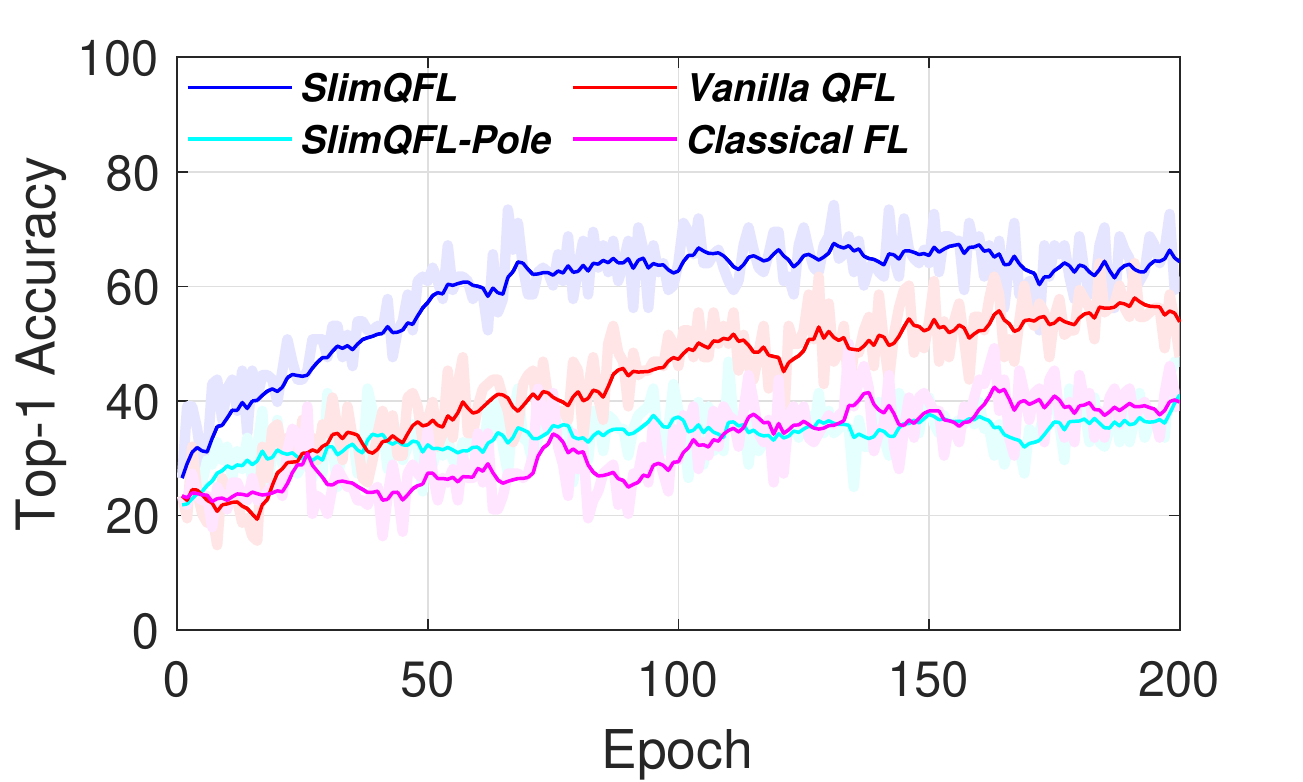}\\
\caption{ Moderate channel ($\sigma^2=-30$dB).}
    \label{fig:result-moderate}
    \end{minipage}\hfill
       \noindent\begin{minipage}{0.33333\textwidth}
     \centering
\includegraphics[width=1\linewidth]{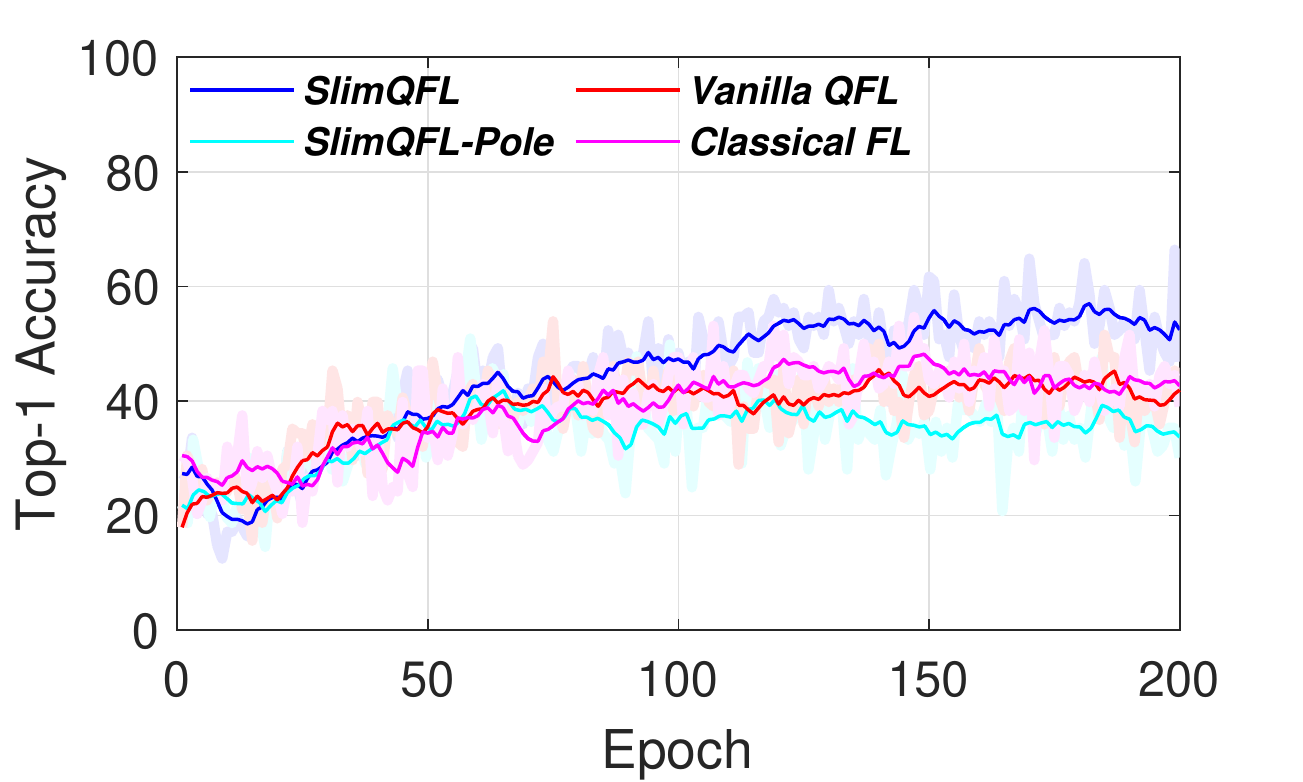}\\
\caption{ Poor channel ($\sigma^2=-20$dB).}
    \label{fig:result-poor}
    \end{minipage}\hfill
\end{figure*}

To show the effectiveness of \textsf{SlimQFL} with both pole training and angle training of QSNNs, we consider the following baselines: \textsf{SlimQFL-Pole} with only pole training of QSNNs, \textsf{Vanilla QFL} with QNNs \cite{chehimi2022quantum}, and \textsf{Classical FL} with classical NNs \cite{mcmahan2017communication}. For a fair comparison, we consider that QSNNs, QNNs, and classical NNs have almost the same number of trainable parameters. The performance is measured using the top-1 accuracy in an MNIST classification task. Since quantum computing suffers from the lack of input qubits, we consider a \textit{mini-version of the MNIST (mini-MNIST)} task, where the images are interpolated into 7x smaller sizes via inter-area interpolation. Compared to MNIST, mini-MNIST consists of 4 labels (\textit{i.e.}, $\{0,1,2,3\}$). We conduct all experiments on mini-MNIST with \textit{independent and identically distributed (IID)} data. 
Other important simulation parameters are summarized in Table~\ref{tab:tab_parameters}. The bold-faced parameters imply the values used for Fig.~\ref{fig:result-good}--Fig.~\ref{fig:result-poor}.

To showcase the dynamic characteristics of SlimQFL, we consider time-varying channel conditions in the uplink communications from each device to the server, while the downlink communications are assumed to be perfect. In the uplink, the throughput $R^n$ of the $n$-th device is given as $R^n =  \log_2(1 + {g^n }/{\sigma^2})$ (bits/sec), where $\sigma^2$ is a constant noise power and $g^n \sim \text{exp}(1)$ is a random gain under Rayleigh fading~\cite{TseBook:FundamaentalsWC:2005}. When the transmitter encodes an input with a code rate $u$, its receiver successfully obtains the encoded data if $R > u$. Then, we consider the standard opportunistic transmission by assuming the channel condition is known at the transmitter. Namely, in a good channel condition, 
the device transmits both pole and angle parameters $[\tilde{\thetab}^n; \tilde{\phis}^n]$ to the server, while in a poor channel condition, 
the device transmits only pole parameters $\tilde \thetab^n$.
Consequently, the server constructs a global QSNN by aggregating the receptions, while counting the pole and angle parameter receptions using $c^n_{\thetab}= \mathbbm{1}{(R^n \geq u^{\textit{pole}}_{\textit{th}})}$ and $c^n_{\phis}= \mathbbm{1}{(R^n \geq u^{\textit{whole}}_{\textit{th}})}$, respectively.

\begin{table}[t!]
    \caption{List of simulation parameters.}
    \tiny
    \centering
    \resizebox{\columnwidth}{!}{
    \begin{tabular}{l|r}
    \toprule[1pt]
      \bf{Description}                & \bf{Value}  \\ \midrule
        Number of devices ($N$)       & \{2,5,\textbf{10},20\} \\
        Number of local iterations per epoch ($L$)       & \{2,5,\textbf{10},20\} \\
        Epoch ($E$)                   & 200 \\
        Optimizer                     & SGD \\
        Learning rate ($\eta_0$), Decaying rate      & $0.01$, $0.001$ \\
        Observable hyperparameter ($w$) & 1.6\\
        Number of qubits               & 4\\
        Number of parameters in \textsf{SlimQFL \& Vanilla QFL}    & 40\\
        Number of parameters in \textsf{SlimQFL-Pole}    & 4\\
        Number of parameters in \textsf{Classical FL} & 56\\
        Number of data per device     & 64\\
        Batch size ($B$)                   & $\{4,8,16,\textbf{32}\}$\\
        Test batch size                   & 128\\
        Noise ($\sigma^2$) & $\{-20, -30, -40\}\,\text{dB}$\\
        \bottomrule[1pt]
    \end{tabular}
    }
    \label{tab:tab_parameters}
\end{table}

\begin{figure*}[t!]
       \noindent\begin{minipage}{0.33333\textwidth}
     \centering
\includegraphics[width=1\linewidth]{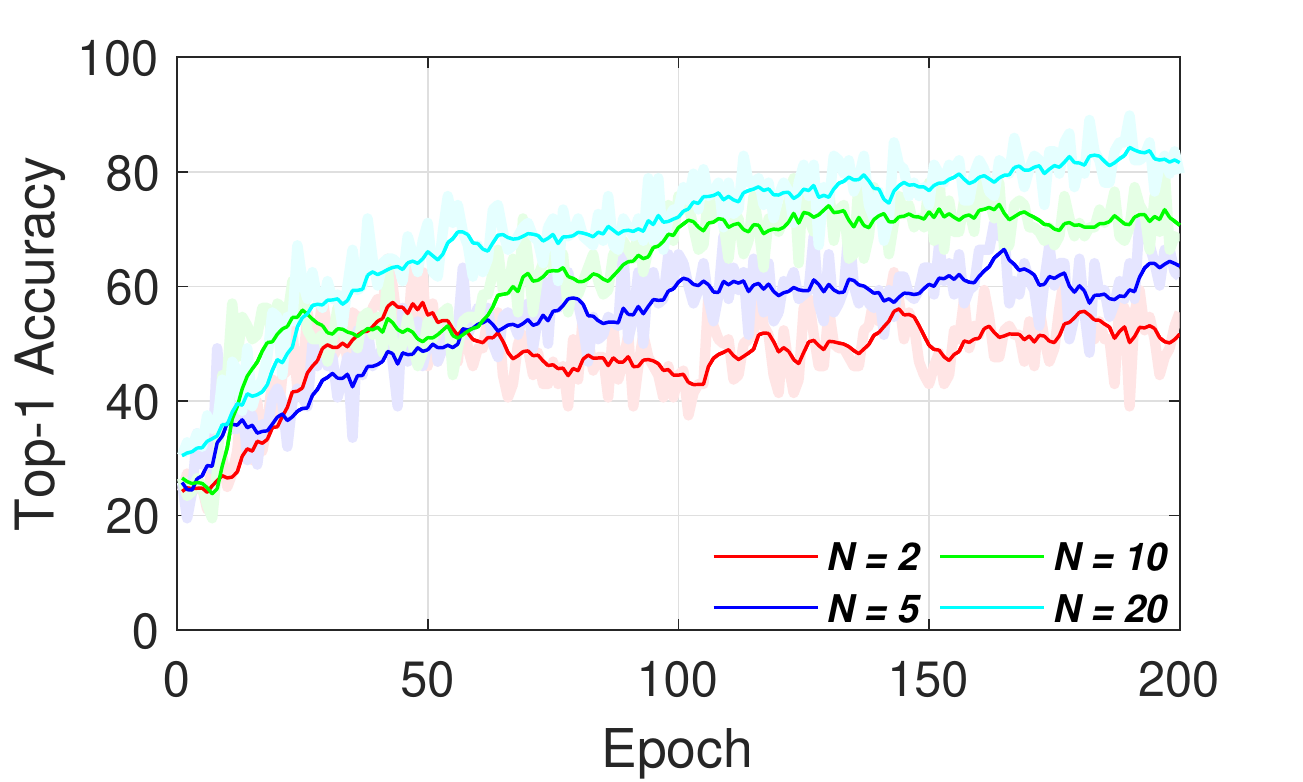}\\
\caption{ Number of devices.}
    \label{fig:result-device}
    \end{minipage}\hfill
       \noindent\begin{minipage}{0.33333\textwidth}
     \centering
\includegraphics[width=1\linewidth]{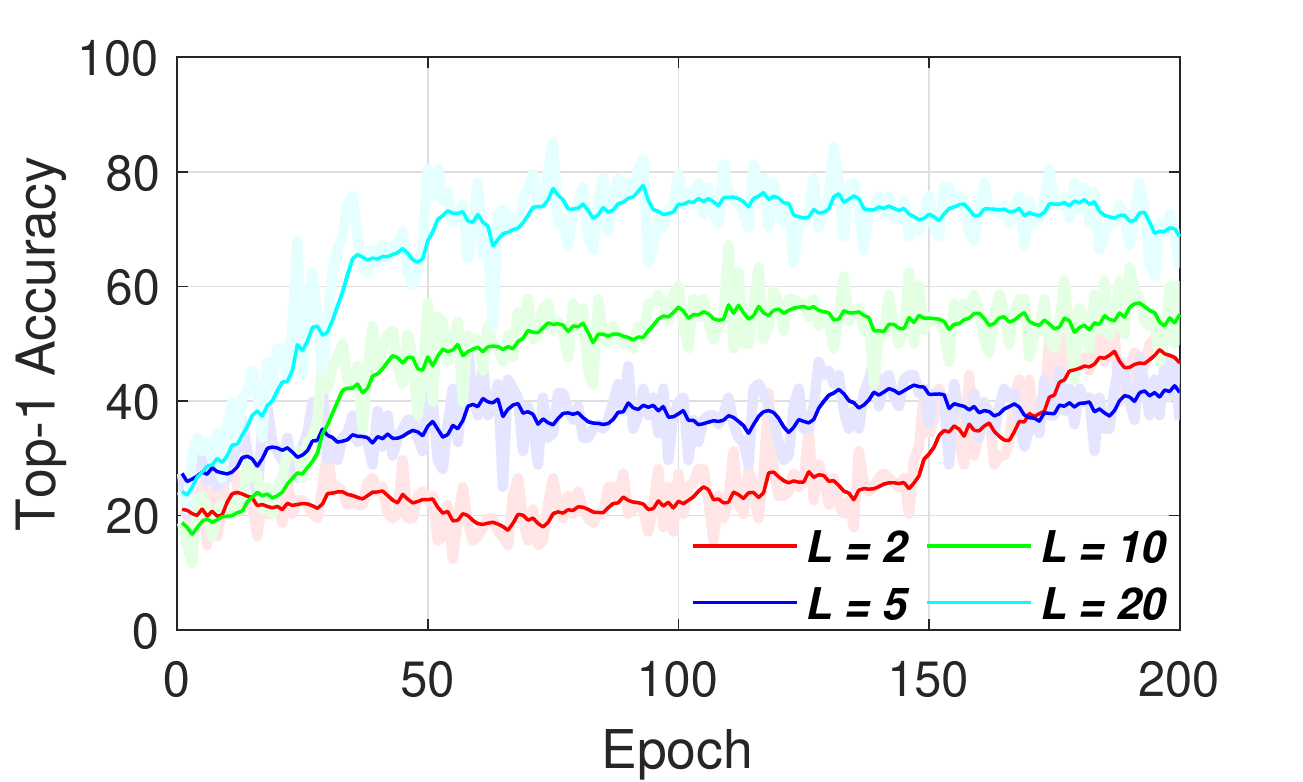}\\
\caption{ Number of local iterations.}
    \label{fig:result-iteration}
    \end{minipage}\hfill
   \noindent\begin{minipage}{0.33333\textwidth}
     \centering
\includegraphics[width=1\linewidth]{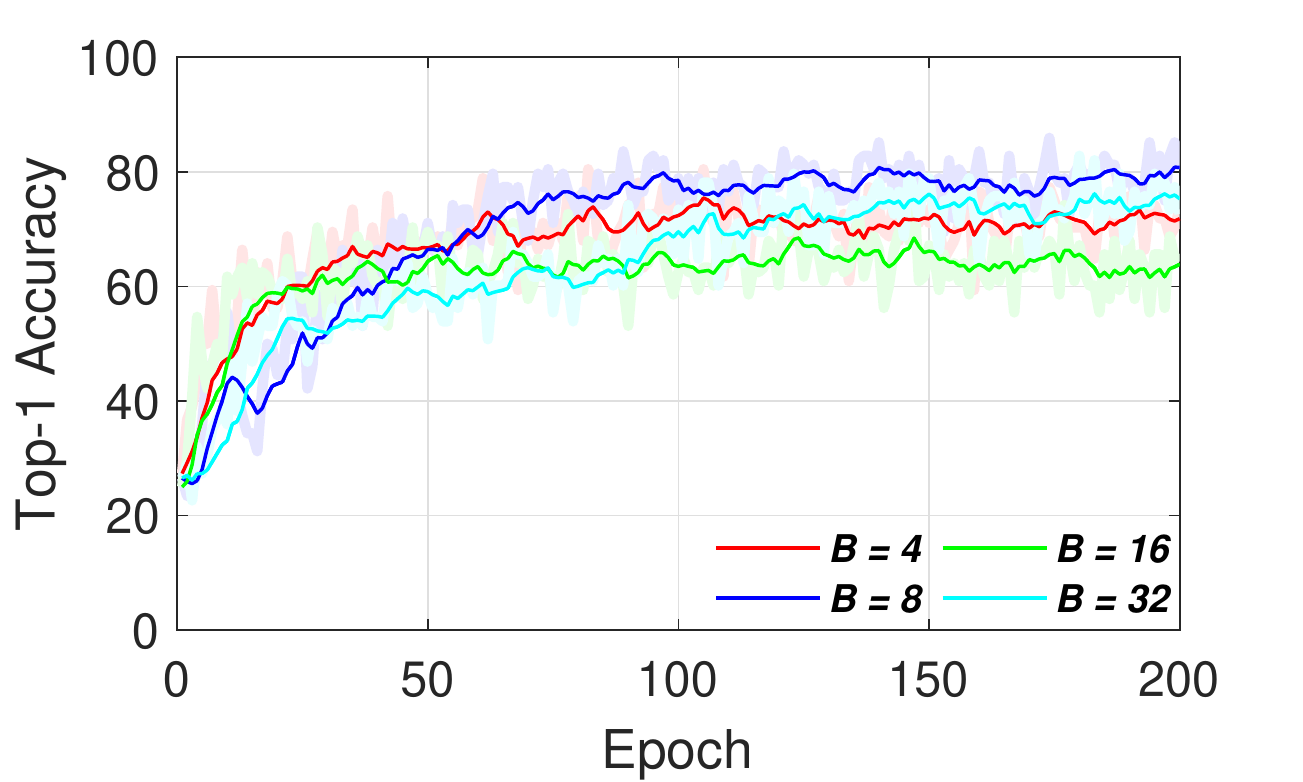}\\
\caption{ Batch size.}
    \label{fig:result-batch}
    \end{minipage}\hfill
\end{figure*}

\subsection{Results} 

\BfPara{Effectiveness of QSNN}
As mentioned above, the proposed scheme changes the parameters being transmitted according to the channel conditions. To examine the robustness of SlimQFL in various channel conditions, we consider three different conditions (i.e., good, moderate, and poor). 
The proposed scheme is expected to produce high accuracy with good channel condition and maintains that result even as the channel condition deteriorates, thanks to the advantage of transmission opportunity. 
Our prediction is verified in Fig.~\ref{fig:result-good}--\ref{fig:result-poor} where we confirm that in good channel conditions, all four schemes yield high accuracy. Despite \textsf{SlimQFL-Pole} having small parameters and low performance, its learning curve shows that \textsf{SlimQFL-Pole} is trained.
 However, as the channel condition deteriorates, the performance degradation of \textsf{Vanilla QFL} and \textsf{Classical FL} is observed while \textsf{SlimQFL-Pole} and \textsf{Proposed} maintain similar levels of accuracy. Among them, \textsf{Proposed} shows the highest accuracy.
Therefore, we conclude that our proposed scheme is superior to the other algorithms.

\BfPara{Effectiveness of Quantum Computing} 
The efficacy of quantum computing is compared with four different schemes.
FL using QNNs (i.e., \textsf{Proposed}, \textsf{Vanilla QFL}) is predicted to show superior performance to its competitors due to the utilization of quantum computing.
This is proven true by the results of Fig.~\ref{fig:result-good} which shows the accuracy results of each algorithm in a good channel condition. \textsf{Proposed} and \textsf{Vanilla QFL} outperforms \textsf{Classical FL} in terms of top-1 accuracy. However, \textsf{SlimQFL-Pole} shows low performance because it consists of only pole parameters.
Thus, we conclude that FL utilizing QNNs has outperformed the classical FL in terms of top-1 accuracy level due to the property of quantum computing. 

\BfPara{Number of Devices}
The number of devices, $N$, affects the model's performance. 
According to \cite{Li2019Convergence}, the more participating devices, the higher the performance of the model.
In Fig.~\ref{fig:result-device}, we see that SlimQFL with 20 devices shows significantly improved results compared to the other baselines which verify the statement above. 

\BfPara{Number of Local Iterations}
Similarly, the number of local iterations, $L$, has a notable impact on the performance. 
According to \cite{Li2019Convergence}, \texttt{FedAvg} shows a higher performance when the number of local iterations is small but when the number of local iterations exceeds a certain threshold, the performance declines (\textit{e.g.,} 20 \cite{wang2020federated}).
It is possible that this trend may not apply to our paper because of QSNN.
Fig.~\ref{fig:result-iteration} corroborates that the larger number of local iterations, the top-1 accuracy also increases. Our results matched the research trends in FL.

\BfPara{Batch Size}
We investigate the impact of the batch size used for training by carrying out simulations using 4, 8, 16, and 32 as batch sizes. Fig.~\ref{fig:result-batch} shows the simulation results. The results in the figure show that the simulation with a batch size of 8 produces the highest accuracy while the simulation with a batch size of 16 has the lowest accuracy. By observing the accuracy of the other two simulations, it is deduced that there is no coherent relationship between model performance and the batch size used in training.

\section{Conclusion}
In this paper, we propose SlimQFL, a novel QFL framework with QSNN and a training algorithm to achieve adaptability to a dynamic communication environment.
The numerical results corroborate that SlimQFL is robust to poor channel conditions compared to QFL or classical FL. Furthermore, we verify that there is a certain advantage in utilizing quantum computing. 
Based on these remarkable results, it could be an interesting topic to analyze the convergence of SlimQFL under various communication channel conditions as well as global data distributions that would be non-IID.

\bibliographystyle{IEEEtran}

\end{document}